\documentclass{article} 
\usepackage{iclr2025_conference,times}

\usepackage{amsmath,amsfonts,bm}

\def\eqref#1{equation~\ref{#1}}

\def\1{\bm{1}}

\DeclareMathAlphabet{\mathsfit}{\encodingdefault}{\sfdefault}{m}{sl}
\SetMathAlphabet{\mathsfit}{bold}{\encodingdefault}{\sfdefault}{bx}{n}

\newcommand{\E}{\mathbb{E}}

\newcommand{\R}{\mathbb{R}}

\usepackage{graphicx}
\usepackage{hyperref}
\usepackage{url}
\usepackage{multirow}
\usepackage{marvosym}

\title{Exploring the Reliability of Self-explanation and its Relationship with Classification in Language Model-driven Financial Analysis}

\author{
Han Yuan \& Li Zhang \& Zheng Ma\textsuperscript{\Letter} \\
Global Decision Science, American Express \\
\texttt{\{Han.Yuan1, Li.Zhang1, Zheng.Ma2\}@aexp.com}
}

\makeatletter
\def\UrlAlphabet{%
      \do\a\do\b\do\c\do\d\do\e\do\f\do\g\do\h\do\i\do\j%
      \do\k\do\l\do\m\do\n\do\o\do\p\do\q\do\r\do\s\do\t%
      \do\u\do\v\do\w\do\x\do\y\do\z\do\A\do\B\do\C\do\D%
      \do\E\do\F\do\G\do\H\do\I\do\J\do\K\do\L\do\M\do\N%
      \do\O\do\P\do\Q\do\R\do\S\do\T\do\U\do\V\do\W\do\X%
      \do\Y\do\Z}
\def\UrlDigits{\do\1\do\2\do\3\do\4\do\5\do\6\do\7\do\8\do\9\do\0}
\g@addto@macro{\UrlBreaks}{\UrlOrds}
\g@addto@macro{\UrlBreaks}{\UrlAlphabet}
\g@addto@macro{\UrlBreaks}{\UrlDigits}
\makeatother
\iclrfinalcopy
\begin{document}

\maketitle

\begin{abstract}
Language models (LMs) have exhibited exceptional versatility in reasoning and in-depth financial analysis through their proprietary information processing capabilities. Previous research focused on evaluating classification performance while often overlooking explainability or pre-conceived that refined explanation corresponds to higher classification accuracy. Using a public dataset in finance domain, we quantitatively evaluated self-explanations by LMs, focusing on their factuality and causality. We identified the statistically significant relationship between the accuracy of classifications and the factuality or causality of self-explanations. Our study built an empirical foundation for approximating classification confidence through self-explanations and for optimizing classification via proprietary reasoning.
\end{abstract}

\section{Introduction}
\footnotetext{\textsuperscript{\Letter} Correspondence: Zheng Ma, Singapore Decision Science Center of Excellence, American Express, 1 Marina Boulevard, 018989, Singapore.}
Recent advances in model architectures, computing hardware, and data resources have positioned language models (LMs) as versatile problem solvers in various domains, and previous studies have also highlighted the potential of LMs in financial classification \citep{lee2024comprehensive,kirtac2024sentiment,arslan2021comparison,xie2024finben,fatemi2025comparative}. 
In formal terms, classification by LMs differs from open-ended generation in that it involves structured generation constrained by a predefined set of options. For instance, in stock trading, classifications are typically limited to two discrete actions: long or short \citep{chuang-yang-2022-buy,koa2024learning,bao2024data}. 
In such cases, LMs are required to provide a single, definitive response rather than ambiguous recommendations, such as suggesting that both options could be reasonable under certain conditions.

Previous research on financial classification has primarily focused on improving the main target: accuracy of classifications \citep{guo-etal-2023-chatgpt,chen-etal-2024-fintextqa,li-etal-2023-chatgpt}. However, explanations of classifications also warrant attention. A classification without accompanying explanation can lead to severe consequences in finance, such as asset losses in the stock trading example. In addition, the absence of explanations makes it challenging for human experts to promptly assess the validity of classifications made by LMs, potentially undermining trust and utility.

Recent advancements such as DeepSeek \citep{liu2024deepseek,lu2024deepseek,liu2024deepseek} show the potential of self-explanations and reasoning to enhance model performance without explicit user instructions. While overall performance gains have been observed, a quantitative evaluation of the relationship between self-explanations and generation quality is essential to establish empirical evidence supporting the reliability of this approach. Some studies have explored the role of proprietary explanations and reasoning in enhancing generation by LMs \citep{zhao-etal-2023-verify}. \citet{lampinen-etal-2022-language} examined the effect of providing a few in-context examples to LMs' prompts and concluded that explanations improved model performance in general domains. Furthermore, \citet{10.5555/3600270.3602472} investigated the use of triplets comprising a question, classification, and explanation in few-shot examples, demonstrating LMs tend to generate nonfactual explanations when making wrong predictions.

\begin{figure}[htbp]
    \centering
\includegraphics[width=\textwidth]{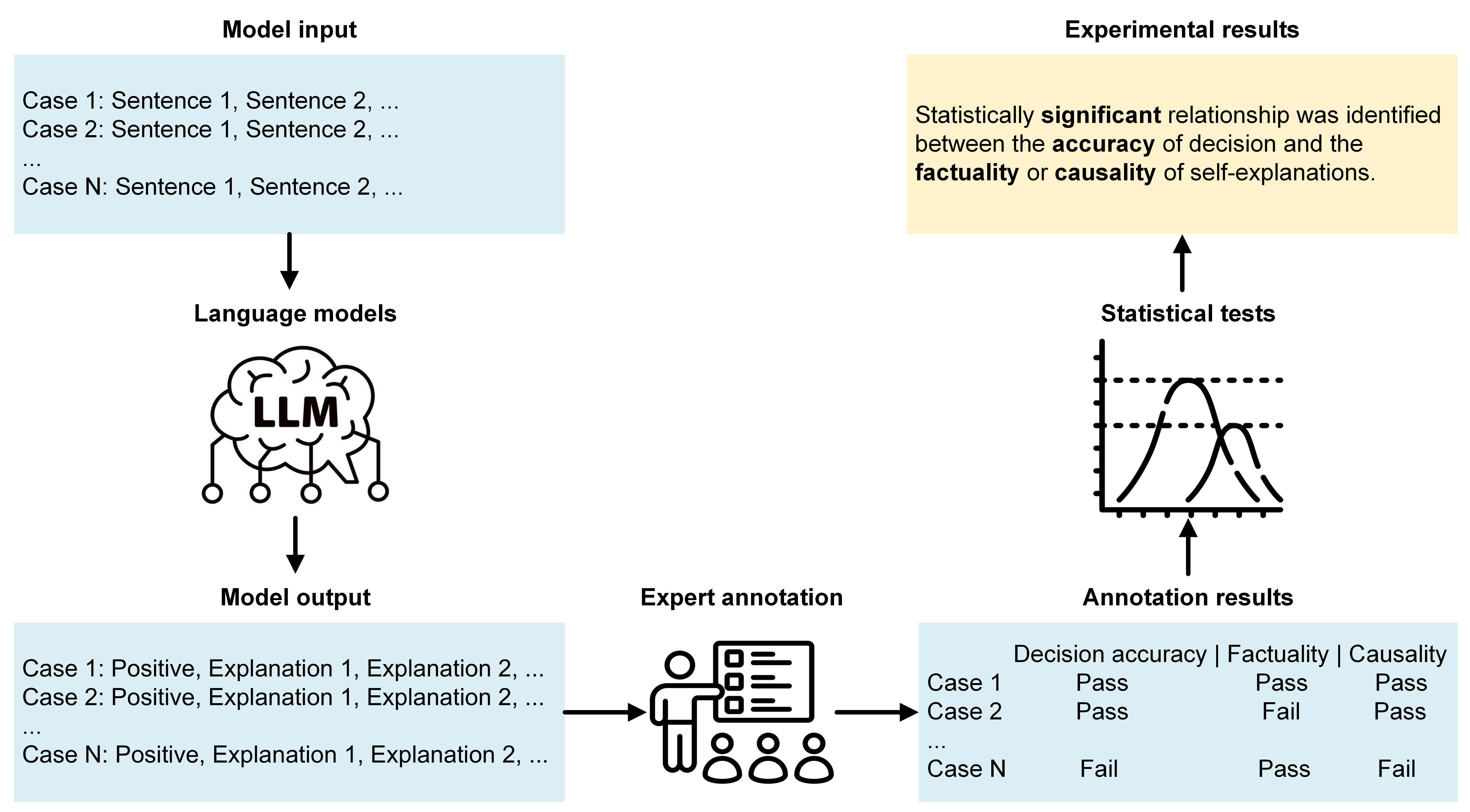}
    \caption{Schematic plot of our experimental pipeline}
    \label{fig:example}
\end{figure}

Our study, as shown in Figure \ref{fig:example} extends this line of research in three key ways. First, we focused on the financial domain, where classification errors can lead to substantial real-world consequences. Second, we investigated zero-shot classification scenarios, where explanations, referred to as self-explanations, were generated by LMs without the aid of in-context examples. Third, we conducted detailed annotations of self-explanations in these scenarios to quantify the relationship between the accuracy of classifications and the factuality or causality of self-explanations, statistically confirming the consistency of observations reported by \citet{10.5555/3600270.3602472} within the financial domain. 

Our experimental finding established an empirical foundation for advancing two research directions. First, it enables the development of a proxy confidence metric of the classification accuracy, such as the proportion of factual or causal inconsistencies in the explanations for a given case. In practical scenarios, such as classifying the hawkish or dovish stance in Federal Open Market Committee (FOMC) speeches, even experts struggle to make highly accurate classifications. Consequently, they cannot reliably assess whether the classifications made by LMs are trustworthy while the factuality and causality of explanations would be more easily to evaluate. With the proxy confidence value, users can make more informed choices regarding the adoption of LMs' classifications. The second area is to provide the empirical foundation for financial classification optimization based on LMs' explanations, which can be interpreted as proprietary reasoning underlying classifications. If no correlation exists between LMs' explanations and their classification accuracy, it becomes difficult to justify the premise that improving model explanations or reasoning would enhance financial classification \citep{reflexion,dou-etal-2024-rest,liu2024deepseek}.

The remainder of this paper is structured as follows: Section \ref{Data} presents the public dataset information and relevant data processing. Section \ref{language} details the LMs and other configurations used in this study. Sections \ref{factual} and \ref{causal} analyze the statistically significant relationship between the accuracy of classifications and the factuality or causality of self-explanations in zero-shot classification by LMs. 
Section \ref{discussion} discusses the limitations of the present study and outlines the planned work informed by current findings. Section \ref{conclusion} concludes the study and provides recommendations for future research. Disclaimers are attached at the end and Appendix \ref{appendix} shows the detailed classification performance.

\section{Experiments}
\subsection{Data\label{Data}}
We conducted the experiments using the publicly available German credit dataset \citep{german_credit_data}, a widely recognized benchmark in financial natural language processing (NLP). To reflect the actual classification capability of LMs, we refined the dataset to increase its signal-to-noise ratio and make it better aligned with LMs' training context. All processing steps were entirely ad hoc and did not involve any analytical operations related to the label, ensuring that performance was not influenced by information leakage. For reproduction, the processed dataset in textual format is released\footnote{\url{https://github.com/Han-Yuan-Med/Language-Models-for-Finance/tree/main/Dataset}} and Appendix \ref{appendix} demonstrates the effectiveness of data processing. Also, all experiments were conducted on all minority cases paired with an equal number of majority cases to ensure that the analyses were not adversely affected by the data imbalance \citep{yuan2022autoscore}.

\subsection{Language models\label{language}}
We utilized three general-purpose LMs: Meta’s Llama-3.2-3B \citep{llama}, Microsoft’s Phi-3.5-3.8B \citep{phi}, and Google’s Gemma-2-2B \cite{gemma}. All LMs were utilized in their instruction-tuned versions, with computations in half-precision due to hardware limitations. The code is publicly available for reproduction\footnote{\url{https://github.com/Han-Yuan-Med/Language-Models-for-Finance/tree/main/Code}}.

\subsection{Factuality\label{factual}}
In financial analysis, beyond answers, it is crucial to provide explanations to meet regulatory requirements and support classification recalibration by financial service providers, thereby mitigating potential adverse impacts. A notable advantage of LMs is their ability to generate user-friendly explanations in natural language. However, the quality of these self-explanations should be rigorously evaluated, as self-explanations containing factual inaccuracies, a prevalent issue of LMs, are of little practical value \citep{ji2023survey}. Moreover, if statistically significant relationships exist between the accuracy of classifications and the presence of fabricated information in self-explanations, factuality could serve as a proxy for assessing the confidence of classifications made by LMs.

The classifications and self-explanations generated by the three LMs were analyzed across top 100 cases, comprising 50 positive and 50 negative cases. Detailed annotations have been released\footnote{\url{https://github.com/Han-Yuan-Med/Language-Models-for-Finance/tree/main/Annotation}}. We quantified the prevalence of factuality issues as the ratio of cases in which the explanations contained any factual inaccuracies. An explanation was classified as having a factuality issue if even a single sentence within it was factually incorrect. Further, we conducted a Chi-squared test to evaluate the independence of factuality issues from the accuracy of model classifications. The null hypothesis posited no relationship between the factuality of the self-explanations and the accuracy of the classifications. The P value is lower than 0.05, suggesting statistically significant dependence between the occurrence of factuality and the accuracy of classification at a confidence level of 95\%. 

Table \ref{factuality} presents an analysis of the relationship between factuality and the accuracy of LMs' classifications. Chi-squared tests revealed statistically significant associations between these factors (P $\leq$ 0.05). We also quantified the prevalence of factuality issues in self-explanations generated by different LMs for both positive and negative cases, observing that Llama-3.2-3B exhibited fewer factuality issues compared to the other two LMs.

\begin{table}[h!]
\centering
\caption{Impact analysis of factuality of self-explanations on the accuracy of classifications}
\label{factuality}
\begin{tabular}{ccccc}
\hline
Language   model              & Case type & Issue   prevalence & Chi-square statistic & P value  \\ \hline
\multirow{2}{*}{Llama-3.2-3B} & Positive        & 0.16               & 23.28                & 3.53e-05 \\& Negative         & 0.20               & 72.40                & 1.30e-15 \\ \hline
\multirow{2}{*}{Phi-3.5-3.8B} & Positive        & 0.26               & 67.28                & 1.63e-14 \\& Negative         & 0.22               & 81.36                & 1.57e-17 \\ \hline
\multirow{2}{*}{Gemma-2-2B}   & Positive        & 0.32               & 57.68                & 1.84e-12 \\& Negative         & 0.16               & 96.24                & 1.00e-20 \\ \hline
\end{tabular}%
\end{table}



\subsection{Causality\label{causal}}
After the factuality check, another critical issue with self-explanations is causality, also referred to as logical inconsistency \citep{10.5555/3600270.3602472}. Specifically, the reasoning within a self-explanation may be contradictory, such as when negative attributes are described as contributing to a positive classification, or vice versa. Similar to factuality, causality is essential for assessing the usability of LMs' self-explanations and serves as a potential confidence measure for the reliability of classifications made by LMs. The same Chi-squared test was computed.

The statistically significant relationship was observed between the causality of self-explanations and the accuracy of classifications across diverse LMs in Table \ref{causality}. Also, compared to factuality issues, causality issues were more prevalent in the self-explanations generated by LMs. Among the evaluated models, Gemma-2-2B demonstrated fewer causality issues than the other two LMs. Besides, in most scenarios, while both factuality and causality exhibited statistically significant relationship with classification accuracy, factuality demonstrated a stronger correlation. This finding suggested that factuality, compared with causality, could serve as a better confidence proxy for classification.

\begin{table}[h!]
\centering
\caption{Impact analysis of causality on the accuracy of model classifications}
\label{causality}
\begin{tabular}{ccccc}
\hline
Language   model              & Case type & Issue prevalence & Chi-square   statistic & P value  \\ \hline
\multirow{2}{*}{Llama-3.2-3B} & Positive  & 0.80       & 22.00                  & 6.52e-05 \\ & Negative  & 0.82       & 72.72                  & 1.12e-15 \\ \hline
\multirow{2}{*}{Phi-3.5-3.8B} & Positive  & 0.52       & 46.68                  & 4.48e-10 \\ & Negative  & 0.68       & 62.96                  & 1.37e-13 \\ \hline
\multirow{2}{*}{Gemma-2-2B}   & Positive  & 0.52       & 46.48                  & 4.48e-10 \\& Negative  & 0.52       & 50.16                  & 7.29e-11 \\ \hline
\end{tabular}%
\end{table}



\section{Discussion\label{discussion}}
This study highlights the statistically significant relationship between the accuracy of classifications and factuality or causality of self-explanations. Due to computational constraints, we did not extend our experiments to LMs such as Llama-3.2-90B. Future research should explore these large-scale models, focusing on comparisons with our current findings in terms of factuality and causality. This would help determine whether the increase in susceptibility to hallucination with larger model parameters also hold in financial applications \citep{rawte-etal-2023-troubling,li-etal-2023-evaluating}. 
Also, our experiments were limited to a single task. Future study should evaluate a broader range of tasks to verify the generalizability of our findings.

Building on extensive validation of the statistically significant relationship between the classification accuracy and the factuality or causality of self-explanations across diverse financial NLP tasks, we scheduled two further studies: (1) to fine-tune an automated detector for accurately identifying factuality or causality issues within self-explanations and test its utility as a confidence proxy for LMs' classifications, and (2) to benchmark existing reasoning-enhancement strategies \citep{chen-etal-2024-measuring,hao-etal-2023-reasoning} based on automatically identified errors  in optimizing financial applications.

\section{Conclusion\label{conclusion}}
In this study, we investigated the use of LMs for financial classification. Beyond final classifications, LMs demonstrated the ability to provide human-interpretable explanations. While challenges related to factuality and causality exist in self-explanations, recent advancements in LMs have substantially mitigated the factuality issue. With the continued evolution of 
reasoning capabilities, LMs hold promise for delivering highly accurate and self-explainable classifications in finance.

\subsubsection*{Disclaimer\label{disclaimer}}
This paper is intended solely for informational purposes and is not a product of or intended to constitute any business practice of American Express. The opinions, findings and conclusions of this paper are those of the authors alone and do not reflect the views of American Express.

\bibliography{iclr2025_conference}
\bibliographystyle{iclr2025_conference}

\appendix
\section{Appendix\label{appendix}}
The prior research \cite{pixiu,fintral} on this dataset overlooked the critical role of data processing in enhancing the signal-to-noise ratio and revealing the true capabilities of LMs. Specifically, the original dataset includes outdated information and pre-existing bias \citep{10.1145/3132847.3132938} that pose challenges for LMs. For instance, certain features are denominated in Deutsche Marks, a currency that has been obsolete for over two decades. Also, LMs often exhibit limited sensitivity to numeric reasoning \citep{mishra-etal-2022-lila}.
To address these issues, we excluded features misaligned with contemporary societal contexts where LMs were developed and converted numeric features into percentile representations through binarization.

For evaluation metrics, we adopted standard evaluation metrics of accuracy and F1 score. In addition, financial classification prioritizes weighted costs, emphasizing the greater consequence of false positive to false negative. As specified in the original dataset documentation \cite{german_credit_data}, the cost associated with a false negative is quantified as 5, while that of a false positive is 1. A lower cost indicates superior performance.

For LMs selection, we reported the results of three LMs in the main text along with two financial domain-specific LMs, FinMA-7B \citep{pixiu} and FinTral-7B \citep{fintral}, which have comparable scales to the general-purpose models, based on their original publications.

Table \ref{lms_performance} shows a consistent improvement in weighted cost across all three LMs when using the processed dataset. For accuracy and F1 score, a substantial enhancement was achieved with Llama-3.2-3B, while performance remained comparable for the other two LMs. Notably, in comparison to the instruction-tuned FinMA-7B and FinTral-7B models on the original German dataset, the zero-shot Llama-3.2-3B demonstrated superior performance, highlighting the effectiveness of data processing and ensuring the analytical quality of LMs' self-explanations.

\begin{table}[h!]
\centering
\caption{LMs performance on original and processed data}
\label{lms_performance}
\begin{tabular}{ccccc}
\hline Model input                      & Language   model & Weighted cost   ↓ & Accuracy ↑ & F1 score ↑ \\ \hline
\multirow{5}{*}{Original data}  & Llama-3.2-3B     & 1185              & 0.53       & 0.34        \\
& Phi-3.5-3.8B     &     304              &      0.51      &   0.67          \\
& Gemma-2-2B       &       322            &        0.51    &     0.67        \\
& FinMA-7B         & -                 & -          & 0.17        \\
 & FinTral-7B       & -                 & 0.61       & -           \\ \hline
\multirow{3}{*}{Processed data} & Llama-3.2-3B     & 306               & 0.68       & 0.74        \\
& Phi-3.5-3.8B     &    295               &     0.51       &     0.67        \\
& Gemma-2-2B       &   296  &  0.51  & 0.67  \\ \hline
\end{tabular}
\end{table}

\end{document}